\documentclass{article} 
\usepackage{lmrl2025_workshop,times}

\usepackage{amsmath,amsfonts,bm}

\def\eqref#1{equation~\ref{#1}}

\def\1{\bm{1}}

\DeclareMathAlphabet{\mathsfit}{\encodingdefault}{\sfdefault}{m}{sl}
\SetMathAlphabet{\mathsfit}{bold}{\encodingdefault}{\sfdefault}{bx}{n}

\usepackage{hyperref}
\usepackage{url}
\usepackage{amsmath,amsfonts}
\usepackage{amsthm}
\usepackage{array}
\usepackage[caption=false,font=normalsize,labelfont=sf,textfont=sf]{subfig}
\usepackage{textcomp}
\usepackage{stfloats}
\usepackage{url}
\usepackage{verbatim}
\usepackage{multirow}
\usepackage{booktabs}       
\usepackage{graphicx}
\usepackage{siunitx}
\usepackage{pdfpages}
\newtheorem{theorem}{Theorem}
\usepackage{algorithm}
\usepackage{algpseudocode}
\usepackage{hyperref}       
\usepackage{cuted}
\usepackage{mathtools}
\usepackage{siunitx}
\usepackage{pdfpages}
\newtheorem{remark}{remark}
\newtheorem{lemma}{Lemma}
\usepackage{wrapfig}
\usepackage{fontawesome} 

\title{GRASP: GRAph-Structured Pyramidal Whole Slide Image Representation}

\author{
\textbf{Ali Khajegili Mirabadi}\textsuperscript{1}\textsuperscript{\href{mailto:ali.mirabadi@ubc.ca}{\faEnvelope}},
Graham Archibald\textsuperscript{1},
Amirali Darbandsari\textsuperscript{1},
Alberto \\ \textbf{Contreras-Sanz}\textsuperscript{1,2},
\textbf{Ramin Ebrahim Nakhli}\textsuperscript{1},
\textbf{Maryam Asadi}\textsuperscript{1},
\textbf{Allen Zhang}\textsuperscript{1},\\
\textbf{C. Blake Gilks}\textsuperscript{2},
\textbf{Peter Black}\textsuperscript{2},
\textbf{Gang Wang}\textsuperscript{3},
\textbf{Hossein Farahani}\textsuperscript{1},
\textbf{Ali Bashashati}\textsuperscript{1}\textsuperscript{\href{mailto:ali.bashashati@ubc.ca}{\faEnvelope}}\\
{\footnotesize \textsuperscript{1}School of Biomedical Engineering \& Department of Pathology and Laboratory Medicine,}\\
{
\textsuperscript{2}Vancouver Prostate Centre, 
\textsuperscript{3}BC Cancer Institute, }\\
{  The University of British Columbia}
}

\lmrlfinalcopy 
\begin{document}
\maketitle

\begin{abstract}
  Cancer subtyping is one of the most challenging tasks in digital pathology, where Multiple Instance Learning (MIL) by processing gigapixel whole slide images (WSIs) has been in the spotlight of recent research. However, MIL approaches do not take advantage of  inter- and intra-magnification information contained in WSIs. In this work, we present GRASP, a novel lightweight graph-structured multi-magnification framework for processing WSIs in digital pathology. Our approach is designed to dynamically emulate the pathologist's behavior in handling WSIs and benefits from the hierarchical structure of WSIs. GRASP, which introduces a convergence-based node aggregation mechanism replacing traditional pooling mechanisms, outperforms state-of-the-art methods by a high margin in terms of balanced accuracy, while being significantly smaller than the closest-performing state-of-the-art models in terms of the number of parameters. Our results show that GRASP is dynamic in finding and consulting with different magnifications for subtyping cancers, is reliable and stable across different hyperparameters, and can generalize when using features from different backbones. The model's behavior has been evaluated by two expert pathologists confirming the interpretability of the model's dynamic. We also provide a theoretical foundation, along with empirical evidence, for our work, explaining how GRASP interacts with different magnifications and nodes in the graph to make predictions. We believe that the strong characteristics yet simple structure of GRASP will encourage the development of interpretable, structure-based designs for WSI representation in digital pathology. Data and code can be found in \footnotesize \url{ https://github.com/AIMLab-UBC/GRASP}
\end{abstract}

\section{Introduction}

Though deep learning has revolutionized computer vision in many fields, digital pathology tasks such as cancer classification remain a complex problem in the domain. For natural images, the task usually relates to assigning a label to an image with an approximate size of 256 × 256 pixels, with the label being clearly visible and well-represented in the image. Gigapixel tissue whole-slide images (WSIs) break this assumption in digital pathology as images exhibit enormous heterogeneity and can be as large as 150,000 × 150,000 pixels. Further, labels are provided at the slide level and may be descriptive of a small region of pixels occupying a minuscule portion of the total image, or they may be descriptive of complex interactions between the substructures within the entire composition of the WSI \cite{laak2017a, yang2015a, glocker2019a}.

Multiple Instance Learning (MIL) has become the prominent approach to address the computational complexity of WSI; however, the majority of methods in the literature focus only on a single level of magnification, usually $20\times$, due to the computational cost of including other magnifications \cite{CLAM, DeepMIL, VarMIL, DGCN, PatchGCN, TransMIL, Zhou_CGC, Guan_2022}. Using this magnification, a set of patches from each WSI are extracted and used as an instance-level representation. This neither captures the biological structure of the data nor does it follow the diagnostic protocols of pathologists. That is to say, WSIs at higher magnifications reveal finer details—such as the structure of the cell nucleus and the intra/extracellular matrix—whereas lower magnifications enable the identification of larger structures like blood vessels, connective tissue, or muscle fibers. Further, these structures are inconsistent from patient to patient, slide to slide, and subtype to subtype \cite{histology}. To capture this variability, pathologists generally use a variety of lenses in their inspection of a tissue sample under the microscope, switching between different magnifications as needed. They generally begin with low magnifications to identify regions of interest for making preliminary decisions before increasing magnifications to confirm or rule out diagnoses \cite{tizhoosh}. 

\begin{wrapfigure}{h}{0.5\textwidth}
    \centering
    \includegraphics[width=0.5\textwidth]{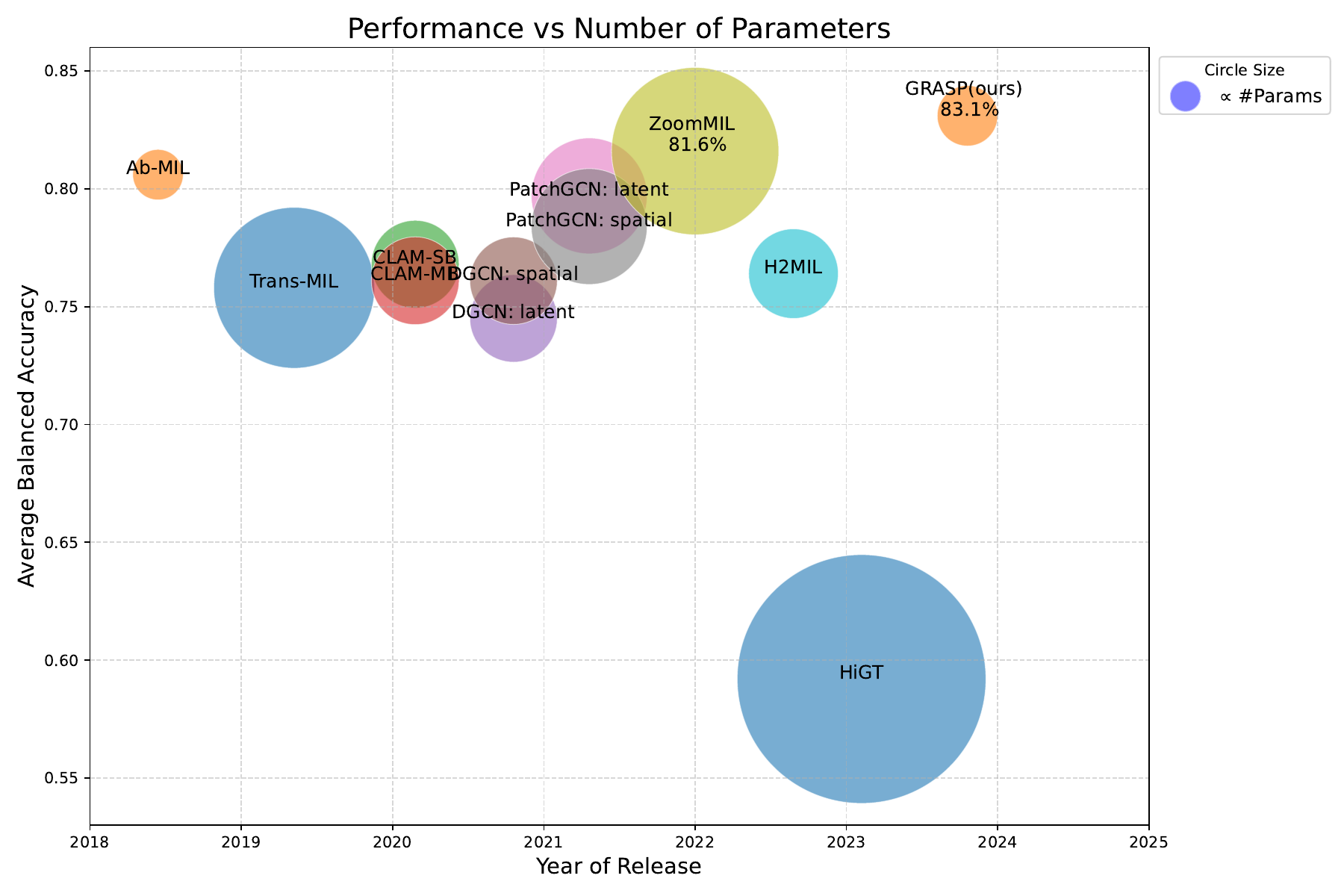}
    \caption{A chronological overview of different WSI representation methods and their performance compared to the size of the model.}
    \label{fig:teaser1}
\end{wrapfigure}

To address this challenge, several multi-magnification approaches have recently been introduced for various tasks such as cancer subtyping, survival analysis, and image retrieval. However, these models often possess millions of parameters, as briefly illustrated in Figure \ref{fig:teaser1}, and suffer from interpretability issues due to their modular complexity \cite{ZoomMIL, DSMIL, KimiaNet, Chen_Vit,multimag-1,multimag-2}. Although these models have demonstrated promise across different tasks, they are not well-suited for low-resource clinical settings, where computational resources are often limited and the infrastructure may not support large-scale computational clusters. Therefore, there is a critical need to develop approaches and models specifically designed for use in smaller clinics, where the hardware may consist of small GPUs with limited memory. These lightweight models must balance accuracy with efficiency, enabling reliable deployment on standard devices while ensuring real-time performance and ease of integration within existing clinical workflows.

In this research, we aim to further the progress of deep learning in this context by introducing a pre-defined fixed structure for a lightweight model that reduces the complexity while maintaining efficacy and interoperability. 
Therefore, our contribution is as follows:
\begin{enumerate}
\itemsep -0.2em 
  \item Introducing GRASP to capture pyramidal information contained in WSIs, as the first \textit{lightweight} multi-magnification model in computational pathology.
  \item GRASP introduces a novel convergence-based mechanism instead of traditional pooling layers to capture intra-magnification information.
  \item We provide a solid theoretical foundation of the model's functionality and its interpretability from both technical and pathological perspectives, as well as providing empirical evidence for the model's efficacy concerning hyperparameters.
  \item An extensive comparison with eleven state-of-the-art models across three different cancers, ranging from two to five histotypes, using two popular backbones demonstrates the generalizability of the proposed method.
\end{enumerate}

 \section{Related Work}
\label{section:2}

\subsection{Patch-Level Encoding} 
With recent progress in deep learning, deep features, i.e. high-level embeddings from a deep network, have advanced past handcrafted features and are considered the most robust sources for image representation. Pre-trained networks such as DenseNet \cite{Densenet}, ResNet \cite{resnet}, or Swin \cite{swin} draw their features from millions of non-medical and non-histopathological images, where they cannot necessarily produce high-level embeddings for complex images, especially rare cancers \cite{Retccl,KimiaNet,martel,ctranspath}.
In this context, the use of Variational Autoencoders (VAEs) has been evaluated in \cite{sish}, where the authors show that DenseNet pre-trained on ImageNet performs better for extracting semantic features from WSIs than VAEs. However, domain-specific vision encoders such as KimiaNet \cite{KimiaNet}, CTransPath \cite{ctranspath}, PLIP \cite{plip}, UNI \cite{uni}, Virchow \cite{virchow}, etc. were developed and trained on large sets of histopathology images (patches) outperforming pre-trained models on ImageNet across various tasks. 

\subsection{Weak Supervision in Gigapixel WSIs}
\textbf{MIL Approaches}:
Several domains of deep learning have been explored in an attempt to effectively address the task of classification in digital pathology. Models such as AB-MIL \cite{DeepMIL}, CLAM \cite{CLAM}, and Trans-MIL \cite{TransMIL} have utilized MIL with promising results. Such approaches have generally focused only on instance-level feature extraction and have not yet explored modeling global, long-range interactions within and across different magnifications. In \cite{mil_surv}, a detailed overview of different MIL methods has been provided.
\\
\textbf{Graph-based Approaches:} 
To incorporate contextual information and long-range interactions, models such as PatchGCN \cite{PatchGCN} and DGCN \cite{DGCN} have been designed with a graph structure that can capture and learn context-aware features from interactions across the WSI. These models represent WSIs as graphs where the nodes are usually embeddings and edges are defined based on clustering or neighborhood node similarity, which in turn adds new hyperparameters and increases inference time. The similarity between nodes can be measured in terms of spatial or latent space, leading to the construction of different graphs for each WSI.
\\
\textbf{Multi-Magnification Approaches}: 
Multiple efforts have been made to incorporate multi-magnification information in the context of gigapixel histopathology subtyping tasks. Models such as HiGT \cite{higt}, ZoomMIL \cite{ZoomMIL}, CSMIL \cite{cross-scale}, $H^2$-MIL \cite{h2mil}, and DSMIL \cite{DSMIL} address this by aggregating contextual tissue information using features from multiple magnifications in WSIs. DSMIL concatenates embeddings from different magnifications together by duplicating lower-magnification features, making the model biased toward lower magnifications and unable to look into inter-magnification information. On the other hand, ZoomMIL aggregates information from $5x$ to $20x$ in a fixed hierarchy with no interaction in the opposite direction. Chen et. al explore this in the context of vision transformers with their Hierarchical Image Pyramid Transformer (HIPT) \cite{Chen_Vit}. Their architecture incorporates regions of size $256 \times 256$ and $4096 \times 4096$ pixels to leverage the natural hierarchical structure of WSIs. 
$H^2$-MIL also adopts a graph-based approach, where it pools the nodes in each magnification using an Iterative Hierarchical Pooling module. Our proposed model, on the other hand, is designed to dynamically aggregate information within and across different magnifications without using traditional pooling layers in its intra-magnification interactions. It also employs a similar mechanism to zoom-in and zoom-out through its inter-magnification interactions, from lower to higher magnifications and vice versa. 


\section{Method}
\label{method}
\label{section:3}

\begin{figure*}[h]
    \centering
    \includegraphics[width=\textwidth]{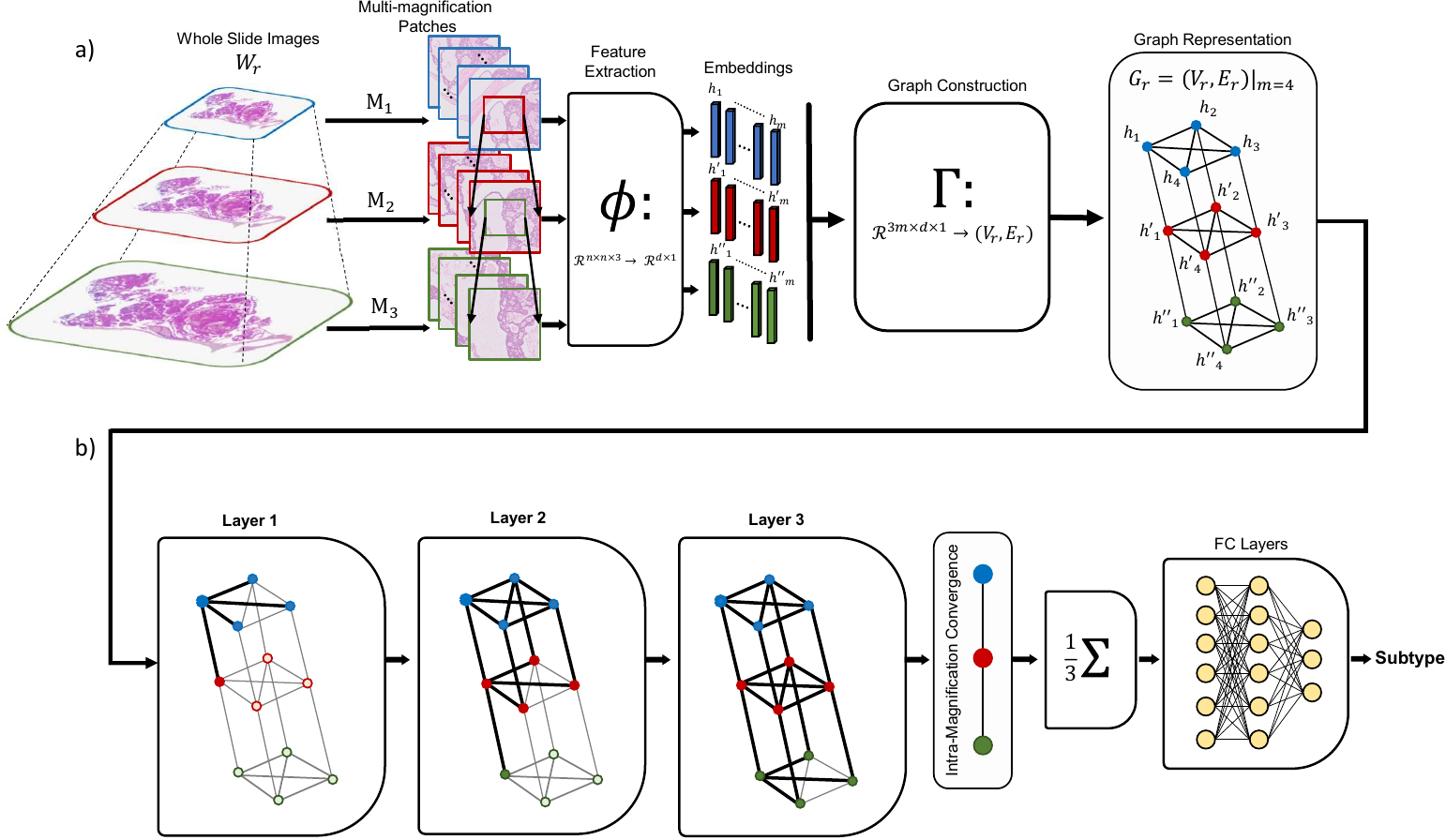}
    \caption{Overview of our workflow beginning with WSIs and outputting slide-level subtype predictions. \textbf{a)} shows the WSI being tiled into patches of varying magnification which are then embedded and assembled into a hierarchical graph. In \textbf{b)}, graph representations are fed into a three-layer GCN  \cite{GCNs} and subsequently, a two-layer MLP to predict graph-level (slide-level) subtypes. As shown in the message passing steps in \textbf{b)}, nodes in the first GCN layer interact with their immediate neighbors; those in the second GCN layer can interact with their second neighbors; and nodes in the final GCN layer can interact with all nodes in the graph. Then, the inter-magnification convergence causes the nodes within each magnification to converge, which is an intrinsic property of the architecture. In the end, the three converged nodes are passed through an average readout module.
    This dynamic helps the model to look for important messages in the entire graph, and if a node contains important information, it will be broadcast to all other nodes in the graph. The output of the GCN layers is then averaged by the \textit{readout module} and passed to the FC layers. (For the sake of illustration, $m=4$ is used to show the structure of GRASP).}
    \label{fig:model}
\end{figure*}
This section introduces the GRAph-Structured Pyramidal (GRASP) WSI Representation, a framework for subtype recognition using multi-magnification weakly-supervised learning, illustrated in Figure \ref{fig:model}.

\subsection{Problem Formulation}
\label{formulation}
Contrary to Multiple Instance Learning (MIL) approaches, which use a bag of instances to represent a given WSI, GRASP benefits from a graph-based, multi-magnification structure to objectively represent connections between different instances across and within different magnifications. To build a graph and learn a graph-based function $\mathcal{F}$ that predicts slide-level labels with no knowledge of patch labels, the following formulation is adopted.

For a given WSI, $W_r \in \mathbb{R}^{N\times M \times 3}$ with label $\mathcal{Y}$, 
three sets of $m$ patches, $\{p_i\in \mathbb{R}^{n\times n \times 3} : \forall  i \in [1,...,m]\}$, 
$\{p^{\prime}_i\in \mathbb{R}^{n\times n \times 3} : \forall  i \in [1,...,m]\}$, and
$\{p^{\prime\prime}_i\in \mathbb{R}^{n\times n \times 3} : \forall  i \in[1,...,m]\}$
are extracted for each magnification of $\textbf{M}_{1}=5x$, $\textbf{M}_{2}=10x$, and $\textbf{M}_{3}=20x$, respectively. 
It is important to note that $p^{\prime\prime}_i$ is the high-resolution window located at the center of $p^{\prime}_i$, and $p^{\prime}_i$ is the high-resolution window located at the center of $p_i$.
\
These patches provide $3m$ patches in total that are then fed into an encoder $\phi$ to encode extracted patches into a lower dimension space as follows:
\begin{equation}
    \phi: p_{i} \longrightarrow h_i \in \mathbb{R}^{d \times 1}, \forall i \in [1,...,m]
    \label{eq1}
    \centering
\end{equation}
where $h_i$ is the feature vector corresponding to the patch $p_i$. Correspondingly, $h^{\prime}_i$ represents $p^{\prime}_i$, and so $h^{\prime\prime}_i$ does $p^{\prime\prime}_i$. Using all the feature vectors for each $W_r$, graph $\mathbb{G}_r$ is constructed using the transformation $\Gamma$:
\begin{equation}
    \Gamma:    
    \left[
    \begin{aligned}
      & \{h_1,...,h_m\}\\
      & \{h^{\prime}_1,...,h^{\prime}_m\}\\
      & \{h^{\prime\prime}_1,...,h^{\prime\prime}_m\}\\
    \end{aligned}  
    \right] \in \mathbb{R}^{3m \times d \times 1}\longrightarrow \mathbb{G}_r = (V_r, E_r) 
    \label{eq2}
    \centering
\end{equation}
Eventually, classifier $\mathcal{C}$ is applied on top of graph convolutional layers $\mathcal{G}$ to build the graph-based function $\mathcal{F}$ to predict slide-level label $\mathcal{Y}$ as follows:
\begin{equation}
    \mathcal{Y} = \mathcal{F}(W_r) = \mathcal{C}(\mathcal{G}(V_r,E_r))
    \label{eq3}
\centering
\end{equation}
\subsection{GRASP}
\label{grasp}
We start by extracting multi-magnification patches as described earlier. Then, for any $i$, we use the same encoder to encode  $p_i$, $p^{\prime}_i$, and $p^{\prime\prime}_i$ into features $h_i$, $h^{\prime}_i$, and $h^{\prime\prime}_i$ respectively. Having instances features, we use the transformation $\Gamma$ to build $\mathbb{G}_r$ as introduced in Eq. \ref{eq2}.

The mechanism of connecting every two nodes in $\mathbb{G}_r$ through $\Gamma$ is premised upon an intuition of the pyramidal nature of WSIs as well as the way in which a conventional light microscope works when one switches from one magnification to another. When using a microscope, increasing magnification preserves the size of the image yet increases resolution by showing the central window of the lower magnification. This is the exact procedure we use to extract our patches in three magnifications. Therefore, for any $i$, $h_i$, $h^{\prime}_i$, and $h^{\prime\prime}_i$ are connected to each other via undirected edges, where this connection represents inter-magnification information contained in the features. On the other hand, for any $i$, all $h_i$'s contain information in $\textbf{M}_{1}$, such that they are connected to each other, forming a fully connected graph at $\textbf{M}_{1}$ magnification to represent intra-magnification information. Similarly, all $h^{\prime}_i$s are connected to each other and also all the $h^{\prime\prime}_i$s to represent intra-magnification information contained in $\textbf{M}_{2}$ and $\textbf{M}_{3}$, respectively.

Figure \ref{fig:model} shows a small example of such a graph for $\mathbb{G}_r = (V_r, E_r) |_{m=4}$ where blue, red, and green nodes each form a fully connected graph of size $m$; the inter-magnification relationship can also be seen via the edges between blue \& red nodes as well as the red \& green nodes.
So far, each WSI, $W_r$, has been represented by a fixed graph $\mathbb{G}_r$ with $3m$ nodes and $\frac{(3m+1)m}{2}$ edges. These graphs are thus deployed to train the GCNs and predict the label $\mathcal{Y}$ at the output.

\subsection{Graph Convolutional Layers}
\label{gcn}
Following Eq. \ref{eq3}, we are defining $\mathcal{G}$ which includes three GCN layers. The intuition behind using three layers is that as a pathologist begins to look for a tumor in a given WSI, they use an initial magnification to find the region of interest; Once found, they consult with other magnifications, which may require zooming in and out back and forth, to confirm their final decision. Therefore, as shown in Figure \ref{fig:model}, all nodes in the graph interact with one another in a hierarchical fashion through the GCN layers. Consequently, each node gradually gathers information from all other nodes, and therefore, if there are any important messages carried by some nodes, it is guaranteed to be broadcast to all other nodes, which is the equivalent of zoom-in and zoom-out mechanism. This dynamic and hierarchical structure imposes theoretical properties on the model which is going to be discussed here. Following the graph convolutional layer introduced in \cite{GCNs}, the graph nodes are updated as follows:
\begin{equation}
    \label{eq4} 
     h_i^{(l+1)} = \alpha(b^{(l)} + \sum_{j\in\mathcal{N}(i)}\frac{1}{c_{ji}}h_j^{(l)}W^{(l)}),
     \centering
\end{equation}
where $b^{(l)}$ is bias; $h_i^{(l+1)}$ is the node feature's update of the graph at $(l+1)$-th step at $\textbf{M}_1$; $\mathcal{N}(i)$ is the set of neighbors of node $i$, $c_{ji} = \sqrt{|\mathcal{N}(j)||\mathcal{N}(i)|}$, where given the symmetry of the graph, all $c_{ji}$s are equal; and $\alpha(.)$ is the activation function, which is ReLU in our implementation. $h_i^{\prime(l+1)}$ and $h_i^{\prime\prime(l+1)}$ expressions follow the same logic as $h_i^{(l+1)}$ in terms of the parameters mentioned above.
\
After the last graph convolutional layer, where the intra-magnification convergence happens, the graph is passed through an average readout module to pool the three-node graph mean embedding and then is fed into the two-layer classifier $\mathcal{C}$ to predict $\mathcal{Y}$.
 \subsection{Intra-Magnification Convergence}
\label{convg}
Based on the idea of capturing the information within and across magnifications, we now show that node features converge in each magnification in the graph to one node Having this, GRASP essentially encodes a graph of size $3m$ nodes to only 3 nodes. We interpret this as the model learning the information contained in the magnification through interaction with other magnifications without a need for traditional pooling layers. 

\begin{theorem}
\label{th:1}
    \textit{Supposing the graph convolutional layers have $L_2$-bounded weights, and the graph node features at $l=0$ are $L_2$-bounded. Therefore, $\forall i, j \in [1,...,m]$, }
    \begin{equation}
        \begin{aligned}
              & \lim_{m\to\infty} ||h_i^{(3)}-h_j^{(3)}||_2=0; \space \lim_{m\to\infty}||h_i^{\prime(3)}-h_j^{\prime(3)}||_2=0; \space 
              \text{and} \lim_{m\to\infty}||h_i^{\prime\prime(3)}-h_j^{\prime\prime(3)}||_2=0.
        \end{aligned} 
    \end{equation}
\end{theorem}
\textit{Proof: Please see Section \ref{sec:IMC} (Theoretical Analysis)}. \\

\textit{Note:} It is worth mentioning that the assumptions made in Theorem \ref{th:1} are minimal. In the case of $L_2$-bounded weights, it has been further explained in \cite{new_oversmooth, oversmoothing}. The assumption that graph node features at $l=0$ are bounded is also minimal as we use a frozen features extractor, which ideally should not generate nondefinitive values for a finite feature vector.\\
With that, we show 
\begin{align}
    & \left\|h_i^{(3)} - h_j^{(3)}\right\|_2 \notag \leq 
    (\frac{1}{m+1})^{3}\left\|h_{i}^{\prime(0)} - h_{j}^{\prime(0)}\right\|_{2}\left\|W^{(2)}\right\|_2\left\|W^{(1)}\right\|_2\left\|W^{(0)}\right\|_2, \notag
\end{align}
which essentially means that $\left\|h_i^{(3)} - h_j^{(3)}\right\|_2$ is upper bounded inversely with the reciprocal of $(m+1)^3$ . Thus, as $m$ increases, the upper-bound gets tighter and eventually leads to  $\lim_{m\to\infty} ||h_i^{(3)}-h_j^{(3)}||_2=0$. As a result, we can conclude the following corollary.

\newtheorem{corollary}{Corollary} 
\begin{corollary}
    \textit{$\forall i\in [1,...,m]$ and $m$ sufficiently large, $h_i^{(3)} \longrightarrow h^{*}$, $h_i^{\prime(3)} \longrightarrow h^{\prime*}$, and $h_i^{\prime\prime(3)} \longrightarrow h^{\prime\prime*}$, where $h^{*}$, $h^{\prime*}$, and $h^{\prime\prime*}$ are functions of $m$; $h^{*}$, $h^{\prime*}$, and $h^{\prime\prime*}$ are the convergence node for each magnification.}
\end{corollary}

This is necessarily equivalent to pooling the nodes in magnification level, yet with a completely new approach than the traditional pooling layer, without further imposing computational load on the network for pooling. Taking into account the fact that $h^{*}$, $h^{\prime*}$, and $h^{\prime\prime*}$ are not necessarily equal, our model is fusing node features in each magnification while it consults with other magnifications, and draw the conclusion via averaging nodes across three magnifications at the end of the convolutional layers by means of the readout module. This means that the final embedding of the graph is $\dfrac{h^{*}+h^{\prime*}+h^{\prime\prime*}}{3}$. We believe that this process helps the model reduce variance and uncertainty in making predictions as $m$ grows. To support this claim, we provide empirical evidence detailed in Section \ref{monte-carlo-test} (Monte Carlo Test).

The structure of the graph has been designed in such a way that it does not get stuck in the bottleneck of over-smoothing, a common issue in deep GCNs \cite{oversmoothing}. Our intuition is that nodes in $\textbf{M}_1$, $\textbf{M}_2$, and $\textbf{M}_3$ are interacting via message passing, and the flow of inter-magnification information helps the model to keep its balance and continue the process of learning. Nevertheless, by increasing the number of GCN layers for the model to four or higher, the so-called over-smoothing problem will take place, which can possibly deteriorate the model's performance. On the other hand, less than three layers of GCNs might not be able to fully capture the inter-magnification interactions. This leaves us with three layers of GCNs, which is equal to the graph's diameter. In addition to this theoretical description, we empirically support our claim in Section \ref{sec:GD} (Graph Depth).

\section{Experiments}
\subsection{Data Preparation}

We utilize three datasets: Esophageal Carcinoma (ESCA) from The Cancer Genome Atlas (TCGA), which includes 135 WSIs across two subtypes, and Ovarian Carcinoma and Bladder Cancer, where Ovarian Carcinoma consists of 948 WSIs with five histotypes, while Bladder Cancer contains 262 WSIs with two histotypes. These datasets were curated using HistoQC \cite{HistoQC}. A detailed breakdown of each dataset is available in Table \ref{table:datasets}.

\subsection{Comparisons with State-of-the-Art}
To compare with state-of-the-art approaches, we repeated these experiments with the same cross-validation folds and random seeds to have a fair comparison; the choice of ten random seeds is to capture statistical significance and reliability. For evaluating the models, we adopt Balanced Accuracy and F1 Score since these metrics show how reliable a model performs on imbalanced data, and more importantly on clinical applications. We compare our proposed model, GRASP, with models using different approaches to have a broad spectrum of evaluation. These models include Ab-MIL \cite{DeepMIL}, Trans-MIL \cite{TransMIL}, CLAM-SB \cite{CLAM}, and CLAM-MB \cite{CLAM} from the attention/transformer-based family; ZoomMIL (2021) \cite{ZoomMIL}, H2MIL (2022) \cite{h2mil}, and HiGT (2023) \cite{higt} from multi-magnification approaches since they have a hierarchical structure and are compatible with our patch extraction paradigm; and PatchGCN: latent \& spatial \cite{PatchGCN} and DGCN: latent \& spatial \cite{DGCN} from graph-based learning approaches.

\subsection{Subtype Prediction}

Table \ref{table:result2} shows the comparison between our model and state-of-the-art methods based on Swin features, where GRASP outperforms all the competing methods on the Ovarian and Bladder datasets. Interestingly, Ab-MIL and CLAMs are the closest-performing methods to GRASP on these two datasets. On the ESCA dataset, however, ZoomMIL is the superior model with GRASP being the closest counterpart. Overall, GRASP is the superior model among all other models based on the average Balanced Accuracy on the three datasets.\\
Table \ref{table:result} shows the comparison between our model and state-of-the-art methods based on KimiaNet features, where GRASP outperforms all the competing methods by a margin of $2.6\%-10.7\%$ Balanced Accuracy on the Ovarian dataset and $0.4\%-10.0\%$ on the Bladder dataset. It is worth mentioning that ZoomMIL is the closest-performing model to GRASP, although ZoomMIL has 7 times more parameters than GRASP. PatchGCN and DGCN are not performing comparably to GRASP, even though they are using spatial information that GRASP does not. This implies that a multi-magnification graph structure can potentially show more capability compared to other state-of-the-art approaches in terms of representing gigapixel WSIs. Moreover, single-magnification approaches are faster in terms of inference time than other approaches, especially CLAM-SB which has the lowest inference time. Inference times (per slide) have been calculated on the same machine for all models. 
\begin{table*}[!ht]
\small
\scriptsize
\centering
\caption{The average performance on 3 folds and 10 random seeds based on Swin's features. The \textbf{best} and \underline{second best} average values are highlighted in \textbf{bold} and \underline{underlined}, respectively.}
\label{table:result2}
\scalebox{0.68}{
\begin{tabular}{c||l|l||cc|cc|cc|cc}
\toprule
\multirow{2}{*}{\textbf{Model}}  & \multirow{2}{*}{\textit{Params.}} & \multirow{2}{*}{\textit{Inference}} & \multicolumn{2}{c|}{\textbf{  Ovarian}: \textit{Five subtypes  }}   & \multicolumn{2}{c|}{\textbf{Bladder}: \textit{Two subtypes}}     & \multicolumn{2}{c|}{\textbf{ESCA}: \textit{Two subtypes}} & \multicolumn{1}{c}{\textbf{Average}} \\ \cline{4-10} 
& & & \multicolumn{1}{c}{\scriptsize Balanced Acc.} & \multicolumn{1}{c|}{\scriptsize F1 Score} & \multicolumn{1}{c}{\scriptsize Balanced Acc.} & \multicolumn{1}{c|}{\scriptsize F1 Score} & \multicolumn{1}{c}{\scriptsize Balanced Acc.} & \multicolumn{1}{c|}{\scriptsize F1 Score} & \multicolumn{1}{c}{\scriptsize Balanced Acc. } \\ \midrule
Trans-MIL      & $2.672$M & $0.019$ sec & $0.297 \pm 0.011$  &  $0.244 \pm 0.011$  &  $0.830 \pm 0.037$ &  $0.819 \pm 0.030$ & $0.626 \pm 0.021$ & $0.611 \pm 0.021$ & $0.584$  \\
Ab-MIL   & $0.263$M & $0.015$ sec &  ${0.643 \pm  0.022}$  &  $0.647 \pm 0.020$ & $0.900 \pm 0.013$ & $0.884 \pm 0.023$ & $0.818 \pm 0.010$ & $0.812 \pm   0.004$ & $0.787$  \\
CLAM-SB    & $0.795$M & $0.014$ sec  &  $0.546 \pm 0.062$  &  $0.550 \pm 0.065$ & $\underline{0.903 \pm 0.051}$ & $\underline{0.902 \pm 0.044}$ & $\underline{0.877 \pm 0.067}$ & $0.861 \pm  0.056$ & $0.775$  \\
CLAM-MB    & $0.796$M & $0.015$ sec &  $0.558  \pm 0.044$   & $0.565 \pm 0.042$ & $\underline{0.903 \pm 0.032}$ & $0.901 \pm 0.028$ & $0.848 \pm 0.055$ & $0.833 \pm   0.051$ & $0.769$ \\
DGCN: \textit{latent}   & $0.790$M & $0.098$ sec  & $0.224  \pm 0.017$ & $0.146 \pm 0.017$  & $0.725 \pm 0.052$ & $0.655 \pm 0.108$ & $0.763 \pm 0.051$ & $0.736 \pm  0.059$ & $0.570$ \\
DGCN: \textit{spatial}   & $0.790$M & $0.086$ sec  & $0.210 \pm 0.011$  & $0.133 \pm 0.012$  & $0.700 \pm 0.044$ &  $0.620 \pm 0.074$ & $0.660 \pm 0.049$ & $0.606 \pm    0.053$ & $0.523$ \\
PatchGCN: \textit{latent}  & $1.385$M & $0.099$ sec & $0.397 \pm 0.039$  & $0.362 \pm 0.047$ & $0.537 \pm 0.011$ & $0.351 \pm 0.052$ & $0.855 \pm 0.076$ & $0.847 \pm   0.076$ & $0.596$  \\
PatchGCN: \textit{spatial}  & $1.385$M & $0.110$ sec  & $0.423 \pm 0.042$ & $0.390 \pm 0.053$ & $0.527 \pm 0.020$ & $0.336 \pm 0.017$  & $0.864 \pm 0.080$ & $0.859 \pm  0.077$ & $0.605$  \\
ZoomMIL   & $2.891$M & $0.024$ sec &   $0.640 \pm 0.018$  & ${0.648 \pm 0.011}$ & $0.899 \pm 0.046$ & $0.895 \pm 0.037$ & $\bold{0.889 \pm 0.037}$  & $\bold{0.895 \pm     0.040}$ & $\underline{0.809}$  \\
 HiGT   & $6.388$M & $0.148$ sec &  $0.251 \pm 0.037$  & $ 0.184 \pm 0.049 $   &  $0.755 \pm 0.041$  &  $0.717 \pm 0.023$   &   $0.760 \pm 0.050$  & $0.744 \pm 0.061$  & $0.588$  \\ 
H2MIL  & $0.829$M & $0.092$ sec &   $\bold{0.671 \pm 0.008}$  & $\bold{0.667 \pm 0.024}$   &  $0.900 \pm 0.054$  &  $0.899 \pm 0.044$  &   $0.854 \pm 0.072$  & $0.845 \pm 0.084$   & $0.808$   \\ \midrule
\textbf{GRASP}   \textit{(ours)}    & $0.378$M & $0.024$ sec &  $\underline{0.669 \pm 0.029}$ & $\underline{0.654 \pm 0.041}$ & $\bold{0.905 \pm 0.058}$ & $\bold{0.906 \pm 0.051}$ & $\underline{0.877 \pm 0.111}$ & $\underline{0.872  \pm   0.112}$ & $\bold{0.817}$ \\ \bottomrule
\end{tabular}
}
\end{table*}

\begin{table*}[t]
\centering
\scriptsize
\caption{The average performance on 3 folds and 10 random seeds based on KimiaNet's features. The \textbf{best} and \underline{second best} average values are highlighted in \textbf{bold} and \underline{underlined}, respectively.}
\label{table:result}
\scalebox{0.8}{
\begin{tabular}{c||l||l||cc|cc|c}
\toprule
\multirow{2}{*}{\textbf{Model}} & \multirow{2}{*}{\textit{Params.}} & \multirow{2}{*}{\textit{Inference}} & \multicolumn{2}{c|}{\textbf{Ovarian}: \textit{Five Subtypes}} & \multicolumn{2}{c|}{\textbf{Bladder}: \textit{Two Subtypes}} & \multirow{1}{*}{\textbf{Model's Average}} \\ \cline{4-8} 
& & & \multicolumn{1}{c}{\scriptsize Balanced Acc.} & \multicolumn{1}{c|}{\scriptsize F1 Score} & \multicolumn{1}{c}{\scriptsize Balanced Acc.} & \multicolumn{1}{c|}{\scriptsize F1 Score} & \multicolumn{1}{c}{\scriptsize Balanced Acc. }  \\ \midrule
Trans-MIL & $2.672M$ & $0.019$ sec & $0.647 \pm 0.007$ & $0.632 \pm 0.005$ & $0.868 \pm 0.023$ & $0.877 \pm 0.013$ & $0.758$ \\
Ab-MIL & $\bold{0.263M}$ & $\underline{0.015}$ sec & $0.692 \pm 0.016$ & $0.680 \pm 0.014$ & $0.919 \pm 0.018$ & $0.922 \pm 0.016$ & $0.806$ \\
CLAM-SB & $0.795M$ & $\bold{0.014}$ sec & $0.627 \pm 0.015$ & $0.623 \pm 0.010$ & $0.908 \pm 0.026$ & $0.911 \pm 0.023$ & $0.768$ \\
CLAM-MB & $0.796M$ & $\underline{0.015}$ sec & $0.620 \pm 0.035$ & $0.609 \pm 0.030$ & $0.901 \pm 0.039$ & $0.906 \pm 0.037$ & $0.761$ \\
DGCN: \textit{latent} & $0.790M$ & $0.098$ sec & $0.654 \pm 0.017$ & $0.652 \pm 0.024$ & $0.835 \pm 0.034$ & $0.841 \pm 0.035$ & $0.745$ \\
DGCN: \textit{spatial} & $0.790M$ & $0.086$ sec & $0.654 \pm 0.009$ & $0.652 \pm 0.009$ & $0.867 \pm 0.015$ & $0.875 \pm 0.007$ & $0.761$ \\
PatchGCN: \textit{latent} & $1.385M$ & $0.099$ sec & $0.683 \pm 0.003$ & $0.675 \pm 0.005$ & $0.911 \pm 0.031$ & $0.919 \pm 0.020$ & $0.797$ \\
PatchGCN: \textit{spatial} & $1.385M$ & $0.110$ sec & $0.672 \pm 0.002$ & $0.662 \pm 0.005$ & $0.896 \pm 0.033$ & $0.905 \pm 0.021$ & $0.784$ \\
ZoomMIL & $2.891M$ & $0.024$ sec & $\underline{0.701 \pm 0.020}$ & $\bold{0.690 \pm 0.021}$ & $\underline{0.931 \pm 0.008}$ & $\underline{0.933 \pm 0.009}$ & $\underline{0.816}$ \\ 
 HiGT   & $6.388$M & $0.148$ sec   &  $0.337\pm0.044$  &  $0.288\pm0.054$   &   $ 0.847\pm  0.067$  & $ 0.842\pm 0.055$  & $0.592$  \\ 
H2MIL  & $0.829$M & $0.092$ sec &   $0.653 \pm  0.018$  &   $ 0.658\pm 0.032$  & $0.876\pm0.054$   & $0.876  \pm 0.048 $  & $0.764$ \\ \midrule
\textbf{GRASP} \textit{(ours)} & $\underline{0.378M}$ & $0.024$ sec & $\bold{0.727 \pm 0.036}$ & $\underline{0.689 \pm 0.040}$ & $\bold{0.935 \pm 0.011}$ & $\mathbf{0.937 \pm 0.014}$ & $\mathbf{0.831}$ \\ \bottomrule
\end{tabular}
}
\end{table*}

Comparing Tables \ref{table:result2} and \ref{table:result}, all the models performed better with KimiaNet embeddings than with Swin embeddings, which is mostly because KimiaNet has domain knowledge and can provide more contextual features than Swin. Furthermore, GRASP, H2MIL, ZoomMIL, Ab-MIL, and CLAMs showcase robust generalization and effective performance even when utilizing features from different backbones, especially with GRASP being the most robust model. \\

Although \cite{cross-scale} has used attention score distribution to show their model is reliable across different magnifications, we want to step further and adopt a similar logic as first introduced in \cite{gradcam} to define the concept of energy of gradients for graph nodes (\textit{Please see Graph-Based Visualization \ref{sec:GBV} in Appendix}). Therefore, for the first time in the field, we show that an AI model such as GRASP can learn the concept of magnification and behave according to the subtype and slide characteristics. To this end, we formulate an experiment to obtain a sense of each magnification's influence on the model which leads to Figure \ref{fig:histogram}. The main takeaway of this experiment is that depending on the subtype, the distribution of referenced magnifications by GRASP is different. From a pathological point of view, this finding fits our knowledge of the biological properties of each subtype. As an example, we conducted a case study on the bladder dataset, where the micropapillary subtype is known to be diagnosed generally in lower magnification owing to its morphological properties and the structure of micropapillary tumors, whereas UCC needs to be examined in higher magnifications due to its cell- and texture-dependent structure.

On the Ovarian dataset, for the subtype ENOC, endometrioids can often be recognized and identified at low power as they tend to have characteristic glandular architecture occupying contiguous, large areas. Low-grade serous carcinomas (LGSC) can be very difficult at low power due to the necessity of confirming low-grade cytology at high power. For CCOC, clear cell carcinomas have characteristic low-power architectural patterns but can also require high-power examinations to exclude high-grade serous carcinoma with clear cell features, meaning that important information is distributed on all magnifications. The other subtypes, MUC and HGSC, may either show pathognomic architectural features at low power or require high-power examination on a case-by-case basis. According to Figure \ref{fig:histogram}, GRASP collects the information from all three magnifications for MUC and CCOC.

Furthermore, to examine whether GRASP understands the biological meaning of the data, i.e., differentiating between tumor vs non-tumor regions, we conduct a visualization experiment to plot the pixel-level heatmap of patches in multiple magnifications as depicted in Figure \ref{fig:heatmap}.
\subsection{Ablation Study}
Here, we design five experiments to evaluate our proposed model. Firstly, a Monte Carlo test on graph size, i.e., the number of nodes to investigate the impact of the number of nodes on the model's performance (Section Monte Carlo Test \ref{monte-carlo-test}). Secondly, analyzing model performance on individual or pairs of magnifications to study the effectiveness of multi-magnification representation (Section Magnification Test \ref{mul-mag}). Thirdly, we investigate the effect of different graph convolution types (Section Graph Convolutions \ref{ab:gcn}). In the fourth experiment, we study how different models perform when all the patches from a WSI are used (Section Patch Number \ref{patch_num}). Lastly, we empirically show that the graph depth of $d= 3$ is the appropriate choice for our design (Section Graph Depth \ref{sec:GD}).

\section{Conclusion}
\label{section:5}
In this work, we developed GRASP, the first \textit{lightweight} multi-magnification framework for processing gigapixel WSIs. GRASP is a \textit{fixed-structure} model that can learn multi-magnification interactions in the data based on the idea of capturing both the inter- and intra-magnification information. This relies on the theoretical property of the model, where it benefits from intra-magnification convergence to pool the nodes rather than conventional pooling layers. GRASP, with its pre-defined fixed structure, has comparably fewer parameters than other state-of-the-art multi-magnification models in the field and outperforms the competing models in terms of average \textit{Balanced Accuracy} over three complex cancer datasets using two different backbones. For the first time in the field, confirmed by two expert genitourinary pathologists, we showed that our model is dynamic in finding and consulting with different magnifications for subtyping two challenging cancers. 
We also evaluated the model's decision-making to show that the model is learning semantics by highlighting tumorous regions in patches.
Furthermore, we not only run extensive experiments to show the model's reliability and stabilization in terms of its different hyperparameters, but we also provide the theoretical foundation of our work to shed light on the dynamics of GRASP in interacting with different nodes and magnifications in the graph.
To conclude, we hope that the strong characteristics of GRASP and its straightforward structure, along with the theoretical basis, will encourage the modeling of lightweight structure-based design in the field of digital pathology for WSI representation.

\newpage

\section{MEANINGFULNESS STATEMENT}
In digital pathology, a meaningful representation of life involves capturing the intricate, multi-scale structures of biological tissues, similar to how pathologists operate. GRASP (GRAph-Structured Pyramidal Whole Slide Image Representation) aids this process by modeling whole slide images as hierarchical graphs that integrate information across different microscopic magnification levels. This method, though lightweight, improves cancer subtyping accuracy and aligns computational analysis with human diagnostic processes, promoting deeper insights into tissue architecture and disease mechanisms. 

\newpage

\section{Appendix}

\begin{figure*}[ht]
    \centering
    \includegraphics[width=\textwidth]{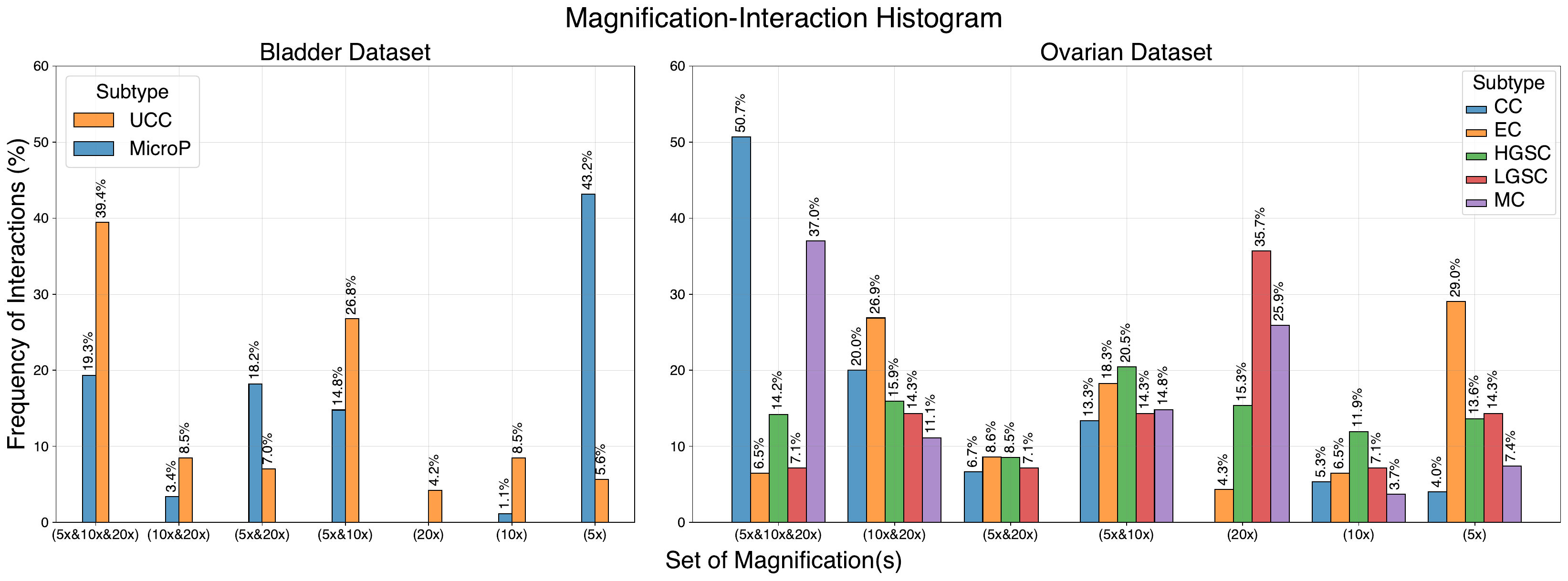}
    \caption{The histogram of consultations conducted by GRASP with different magnifications. First, this shows GRASP is actively dynamic in terms of capturing information from different magnifications benefiting from its multi-magnification structure. Second, information is distributed differently over magnifications depending on the subtype and slide, and there is no optimal magnification for a subtype. For example, in the Bladder dataset, \lq $(5x\&10x\&20x)$' shows that the model needed to consult with all three magnifications for $19.3\%$ and $39.4\%$ of slides for MicroP and UCC, respectively; \lq $(5x)$' shows that the model has mostly focused on only $5x$ magnification for $43.2\%$ and  $5.6\%$ of slides for MicroP and UCC, respectively. This behavior is similar to pathologists, where they can diagnose massive MicroP tumors with lower magnifications, while they need to consult with higher magnifications to confirm a minuscule mass of MicroP tumors. On the other hand, UCC is hard to diagnose at lower magnifications and requires careful examination with different magnifications due to its morphological complexity, which fits the model behavior in proclivity to highlight more than one magnification for the majority of cases.}
    \label{fig:histogram}
\end{figure*}

\begin{figure*}[!ht]
    \centering
    \includegraphics[width=10cm]{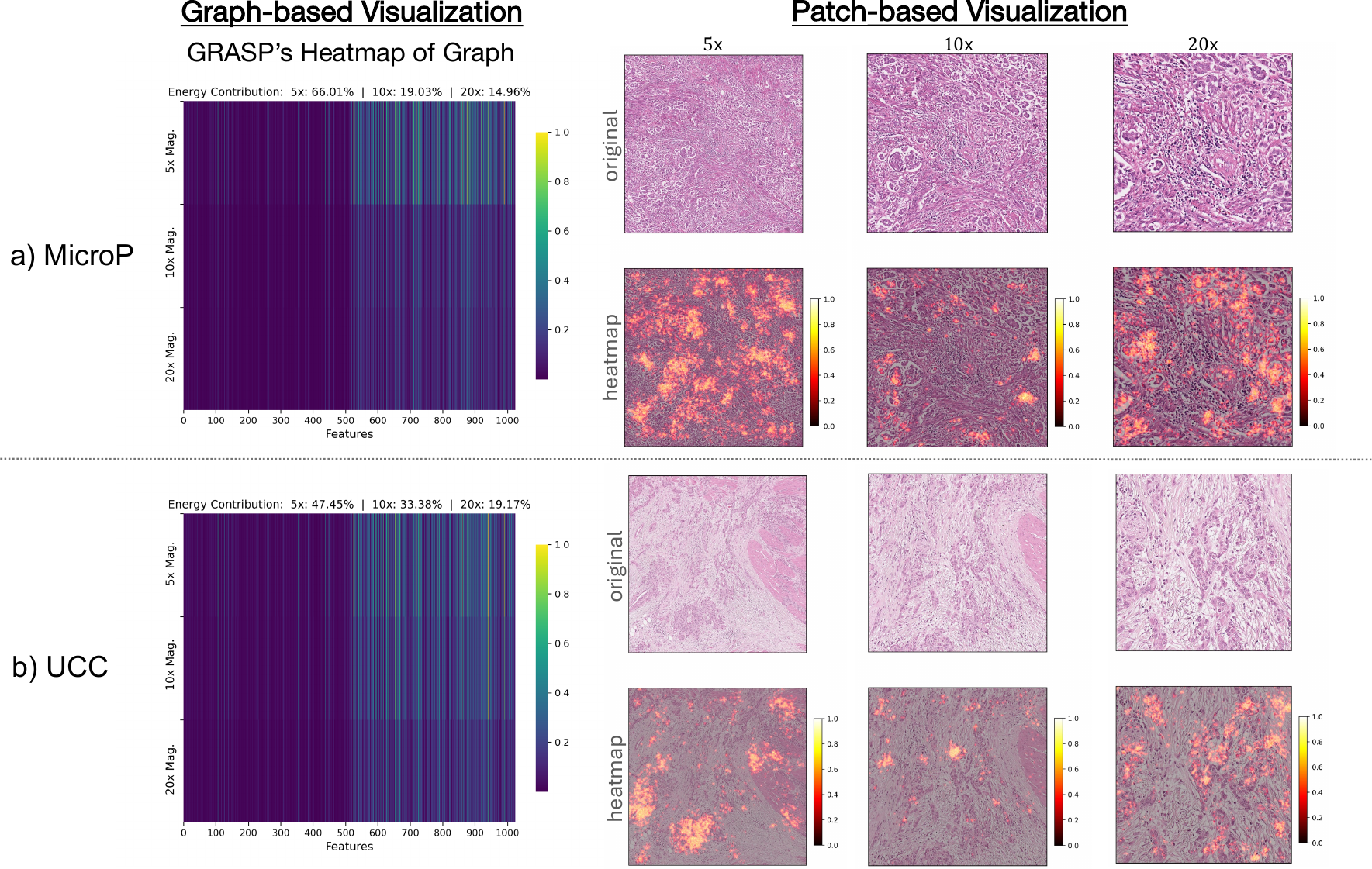}
    \caption{A case study on the Bladder dataset using KimiaNet features. \textbf{a)} Graph-based visualization: a random case from the subtype MicroP in the test data was selected to visualize its magnification heatmap where we show the absolute gradient in terms of each node. The $5x$ magnification contributes to $66.01\%$ of the whole energy model spent on this slide, meaning GRASP overall emphasizes more on $5x$ on this slide. Patch-based visualization: GRASP highlights patches of the three magnifications of a region of interest. In the second row, highlighted regions show the model has identified those areas as important while paying minimal attention to other regions. As confirmed by an expert pathologist, the model's highlights on the three patches are tumors. The model can thus differentiate MicroP tumors from other tissue textures despite being trained for separating MicroP vs UCC. \textbf{b)} shows a similar case yet on the subtype UCC from a random slide in the test data. In this case, GRASP focuses on both $5x (47.45\%)$ and $10x (33.38\%)$ but is more interested in $5x$. As confirmed by the expert pathologist, the regions highlighted (yellowish areas in the second row) by the model are tumorous neighborhoods. Therefore, GRASP can differentiate UCC tumors from other textures and healthy cells across multiple magnifications.}
    \label{fig:heatmap}
\end{figure*}

\subsection{Datasets}

\begin{table*}[h!]
\caption{Summary of the datasets used in this study.}
\centering
\resizebox{\textwidth}{!}{
\begin{tabular}{lccc}
\toprule
\textbf{Dataset} & \textbf{Source} & \textbf{No. of WSIs} & \textbf{Histotypes/Subtypes} \\ \midrule
\multirow{5}{*}{Ovarian Carcinoma} & \multirow{5}{*}{Private Dataset} & \multirow{5}{*}{948} & 
High-Grade Serous Carcinoma (HGSC): 410 \\ \cline{4-4} 
 &  &  & Clear Cell Ovarian Carcinoma (CCOC): 167 \\ \cline{4-4} 
 &  &  & Endometrioid Carcinoma (ENOC): 237 \\ \cline{4-4} 
 &  &  & Low-Grade Serous Carcinoma (LGSC): 69 \\ \cline{4-4} 
 &  &  & Mucinous Carcinoma (MUC): 65 \\ \midrule
\multirow{2}{*}{Esophageal Carcinoma (ESCA)} & \multirow{2}{*}{TCGA} & \multirow{2}{*}{135} & 
Adenocarcinoma: 86 \\ \cline{4-4} 
 &  &  & Squamous Cell Carcinoma: 49 \\ \midrule
\multirow{2}{*}{Bladder Cancer} & \multirow{2}{*}{Private Dataset} & \multirow{2}{*}{262} & 
Micropapillary (MicroP): 128 \\ \cline{4-4} 
 &  &  & Conventional Urothelial Carcinomas (UCC): 134 \\ \bottomrule
\end{tabular}
}
\label{table:datasets}
\end{table*}

A total of 1,133,388 patches of size $1000\times1000$ pixels for the Ovarian dataset, 602,874 patches of size $224\times224$ from the ESCA dataset, and 313,191 patches of size $1000\times1000$ pixels for the Bladder dataset are extracted in multi-magnification setting (approximately 2 TB of Gigapixel WSIs). Patches being extracted such that they do not overlap at $\textbf{M}_3$ while overlapping at $\textbf{M}_2$ and $\textbf{M}_1$ is inevitable. From each magnification, $m \leq 400$ patches (note that we use a large field of view, meaning this number eventually covers much of tissue regions) have been extracted per slide in both Ovarian and Bladder datasets, as it's been shown in \cite{sish,Retccl,tizhoosh} that a subset of patches is enough to represent WSIs. \\
For patch-level feature extraction, we utilized two backbones: KimiaNet and Swin\_base. Given that KimiaNet was trained on TCGA data in a supervised fashion, we intentionally refrained from extracting features from the ESCA dataset using this backbone to ensure an unbiased and leakage-free comparison. Conversely, we employed Swin, pre-trained on ImageNet, to extract features from all three datasets. \\
For each cancer dataset, we trained our proposed method in a 3-fold cross-validation and repeated the experiments \textit{ten} times with different random seeds, where random seeds were randomly generated, to ensure a rigorous comparison. In order to prevent data leakage in our cross-validation splits, we split the slides based on their patients, since some patients have more than one slide, meaning that slides were split in a way that all slides from the same patient remain in the same set.

\subsection{Training and Inference}
To tackle the data imbalance problem, for all models in the study, we deployed a weighted cross entropy loss. A learning rate of $0.001$ and a weight decay of $0.01$ for Adam optimizer have been adopted, and in case competing models were not converging, learning rate of $0.0001$ resolved the problem. Models were trained for 100, 50, and 10 epochs for the Ovarian, ESCA, and Bladder datasets, respectively. Specific to GRASP, the first two layers are of size 256 and the last layer output is of size 128. For all training and testing, the GPU hardware used was either a GeForce GTX 3090 Ti-24 GB (Nvidia), Quadro RTX 5000-16 GB (Nvidia), RTX 6000-48 GB (Nvidia) based on availability. Deep Graph Library (DGL), PyTorch, NumPy, SciPy, PyGeometric, and Scikit-Learn libraries have been used to perform the experiments.

\subsection{Theoretical Analysis}
\label{sec:IMC}
Here, we prove Theorem \ref{th:1} for any $h_{i}^{(3)}$ and $h_{j}^{(3)}$, and conclude the case for $h_{i}^{\prime(3)}$s and $h_{i}^{\prime\prime(3)}$s similarly. To start with, we demonstrate Lemma \ref{lem:1}. 

\begin{lemma}
    \label{lem:1}
    For any given vectors $x$ and $y$, and having $\left\| . \right\|_2$ as the $L_2$ norm, the following inequality holds, \\
    \begin{equation}
        \centering
        \left\|\alpha(x)-\alpha(y)\right\|_2 \leq \left\|x-y\right\|_2, 
    \end{equation}
    \label{eq:6}

    where $\alpha(.) = \operatorname{ReLU}(.)$
\end{lemma}

\begin{proof}
Let's reformulate $\operatorname{ReLU}(x)$ as $\frac{x+\left|x\right|}{2}$ where operator $|x|$ is the element-wise absolute value of the vector $x$. Thus,
        \begin{align}
        & \left\|\alpha(x)-\alpha(y)\right\|_2 = \left\|\frac{x+|x|}{2}-\frac{y+|y|}{2}\right\|_2  \notag \\
        & = \left\|\frac{x-y}{2}+\frac{|x|-|y|}{2}\right\|_2  \notag \\
        & \leq \left\|\frac{x-y}{2}\right\|_2+\left\|\frac{|x|-|y|}{2}\right\|_2 
        \end{align}
    using the reverse triangle inequality, $\left\|\frac{|x|-|y|}{2}\right\|_2 \leq \left\|\frac{x-y}{2}\right\|_2$, which gives result to
        \begin{align}
        \centering
             &  \left\| \alpha(x)-\alpha(y)\right\|_2 \leq \left\|\frac{x-y}{2}\right\|_2+\left\|\frac{x-y}{2}\right\|_2 \notag \\
             & = \left\|x-y\right\|_2 
        \end{align}
\end{proof}

\begin{figure}[h]
    \centering
    \includegraphics[width=3cm]{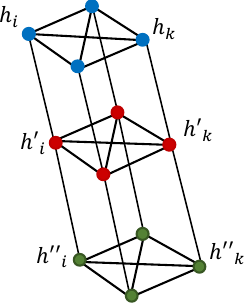}
    \caption{The structure of our hierarchical graph and the relationship between two given nodes $h_i$ and $h_k$ within and across different magnifications.}
    \label{fig:theorem}
\end{figure}


\textbf{Theorem 1:}
\begin{proof}
Recalling the main GCN formula for any $l$ in \ref{eq4},
$\mathcal{N}(i)$ is the neighborhood size, and taking self-loops (\textit{see Note \ref{note}}) into account, $\forall i \in [1,...,m]$, $\left|\mathcal{N}(i)\right|=m+1$. With the graph being symmetric, we deduce that $\forall j \in [1,...,m]$, $\left|\mathcal{N}(j)\right|=m+1$. These result in 
    $$
    c_{ji} = \sqrt{\left|\mathcal{N}(j)\right|\left|\mathcal{N}(i)\right|}=m+1
    $$ hence, Eq. \ref{eq4} is simplified as follows,
    \begin{equation}
        \label{eq:9}
         h_i^{(l+1)} = \alpha(b^{(l)} +\frac{1}{m+1}\sum_{j\in\mathcal{N}(i)}h_j^{(l)}W^{(l)}).
    \end{equation}
Now, for any $i$, we can partition the set of all nodes in $\mathcal{N}(i)$ into two partitions $\{h_1^{(l)},...,h_{m}^{(l)}\}$ and $\{h_{i}^{\prime(l)}\}$ based on the relationship between nodes as shown in Figure \ref{fig:theorem}. Therefore, $\sum_{j\in\mathcal{N}(i)}h_j^{(l)}$ can be rewritten as,
    \begin{equation}
        \label{eq:10}
         \sum_{j\in\mathcal{N}(i)}h_j^{(l)} = \left( \sum_{j\in [1,...,m]}h_j^{(l)} \right)
         + h_{i}^{\prime(l)}.
    \end{equation}
The first term in Eq. \ref{eq:10} is common among all nodes at a given magnification, so we call it $\mathcal{H}^{(l)}$ leading to Eq. \ref{eq:11},
    \begin{equation}
        \label{eq:11}
         \sum_{j\in\mathcal{N}(i)}h_j^{(l)} = \mathcal{H}^{(l)} + h_{i}^{\prime(l)}
    \end{equation}
    as a result, we combine Eq. \ref{eq:9} with Eq. \ref{eq:11} which yields
    \begin{equation}
        \label{eq:12}
         h_i^{(l+1)} = \alpha(b^{(l)} +\frac{1}{m+1} \left( \mathcal{H}^{(l)} + h_{i}^{\prime(l)} \right)W^{(l)})
    \end{equation}
and similarly for any $j \neq i$ as well,
    \begin{equation}
        \label{eq:13}
        h_j^{(l+1)} = \alpha(b^{(l)} +\frac{1}{m+1} \left( \mathcal{H}^{(l)} + h_{j}^{\prime(l)} \right)W^{(l)}).
    \end{equation}
By using Lemma \ref{lem:1} and with combination with Eq. \ref{eq:12} and \ref{eq:13},
    \begin{align}
        \label{eq:14}
        & \left\|h_i^{(l+1)} - h_j^{(l+1)}\right\|_2 \notag \\
        & \leq \left\|\frac{1}{m+1}h_{i}^{\prime(l)}W^{(l)} - \frac{1}{m+1}h_{j}^{\prime(l)}W^{(l)}\right\|_2 \notag \\
        & = \frac{1}{m+1}\left\|\left(h_{i}^{\prime(l)} - h_{j}^{\prime(l)}\right)W^{(l)}\right\|_2 \notag \\
        & \leq \frac{1}{m+1}\left\|h_{i}^{\prime(l)} - h_{j}^{\prime(l)}\right\|_{2} \left\|W^{(l)}\right\|_2.
    \end{align}
Therefore, we reach the inequality below,
    \begin{equation}
        \label{eq:15}
         \left\|h_i^{(l+1)} - h_j^{(l+1)}\right\|_2 \leq \frac{1}{m+1}\left\|h_{i}^{\prime(l)} -
         h_{j}^{\prime(l)}\right\|_{2} \left\|W^{(l)}\right\|_2.\\
    \end{equation}
Now, by going recursively over $l=0,1,2$, we have
\begin{align}
    \label{eq:16}
    & \left\|h_i^{(3)} - h_j^{(3)}\right\|_2 \notag \leq \\
    & (\frac{1}{m+1})^{3}\left\|h_{i}^{\prime(0)} - h_{j}^{\prime(0)}\right\|_{2}\left\|W^{(2)}\right\|_2\left\|W^{(1)}\right\|_2\left\|W^{(0)}\right\|_2.
\end{align}
Since $\left\|W^{(2)}\right\|_2$, $\left\|W^{(1)}\right\|_2$, and $\left\|W^{(0)}\right\|_2$ are $L_2-bounded$ based on our assumption. Also, $\left\|h_{i}^{\prime(0)} - h_{j}^{\prime(0)}\right\|_{2}$ is an $L_2-bounded$ value based on our assumption (\textit{as input image data is $L_2-bounded$, and also our encoder $\phi$ is a bounded encoder: features are not scattered in an infinite space, rather they are encoded in a finite space}).
Given these, by approaching $m\rightarrow\infty$ (\textit{see remark \ref{note2}}), the right side of the Eq. \ref{eq:16} approaches $0$. Therefore,
\begin{equation}
    \label{eq:17}
    \lim_{m\to\infty} \left\|h_i^{(3)}-h_j^{(3)}\right\|_2=0
\end{equation}
similar to this case, it can be proved that
\begin{equation}
    \label{eq:18}
 \lim_{m\to\infty} \left\|h_i^{\prime(3)}-h_j^{\prime(3)}\right\|_2=0 
\end{equation}
\begin{equation}
\label{eq:19}
    \lim_{m\to\infty} \left\|h_i^{\prime\prime(3)}-h_j^{\prime\prime(3)}\right\|_2=0 
\end{equation}
\end{proof}
\begin{center}
    \hfill $\square$ \notag
\end{center}

\begin{remark}
\label{note}
\textit{To implement GCNs, self-loops are considered to represent the relationship between each node with itself, and it is also part of the technical implementation of the models. Thus, we consider this fact in our theoretical discussion}.
\end{remark}

\begin{remark}
 \label{note2}
 \textit{Empirically, reaching sufficiently large $m$ can guarantee the convergence. For example, $m=10$ can affect the upper bound in Eq. \ref{eq:16} with an order of $\frac{1}{10^3}$, while $m=100$ can affect the upper bound with an order of $\frac{1}{10^6}$. In our experiments, $m=400$ has been adopted that guarantees the convergence with an order of $\frac{1}{64\times10^6}$. Therefore, the larger $m$, the tighter together node features at the last GCN layer.}
\end{remark}

\begin{corollary}
    $\forall i\in [1,...,m]$ and $m$ sufficiently large, $h_i^{(3)} \longrightarrow h^{*}$, $h_i^{\prime(3)} \longrightarrow h^{\prime*}$, and $h_i^{\prime\prime(3)} \longrightarrow h^{\prime\prime*}$, where $h^{*}$, $h^{\prime*}$, and $h^{\prime\prime*}$ are functions of $m$; $h^{*}$, $h^{\prime*}$, and $h^{\prime\prime*}$ are the convergence node for each magnification.
\end{corollary}
\textbf{Description}: \textit{given Eq. \ref{eq:17}, every two arbitrary nodes $h_i^{(3)}$ and $h_j^{(3)}$ in one level of magnification are converging to each other. This means that all nodes are converging to the same value, which we name $h^*$. Thus, $\forall i \in [1,...,m]$, $ \lim_{m\to\infty} \left\|h_i^{(3)}-h^*\right\|_2=0$ or equivalently  $h_i^{(3)} \longrightarrow h^{*}$. Using the same logic as above, one can conclude $h_i^{\prime(3)} \longrightarrow h^{\prime*}$ and $h_i^{\prime\prime(3)} \longrightarrow h^{\prime\prime*}$. Since each of $h^*$, $h^{\prime*}$, and $h^{\prime\prime*}$ are a function of $m$, increasing $m$ would result in them being a better estimation/representation for the intra-magnification information.}


\subsection{Empirical Proof}
\begin{figure}[h]
    \centering
    \includegraphics[width=8cm]{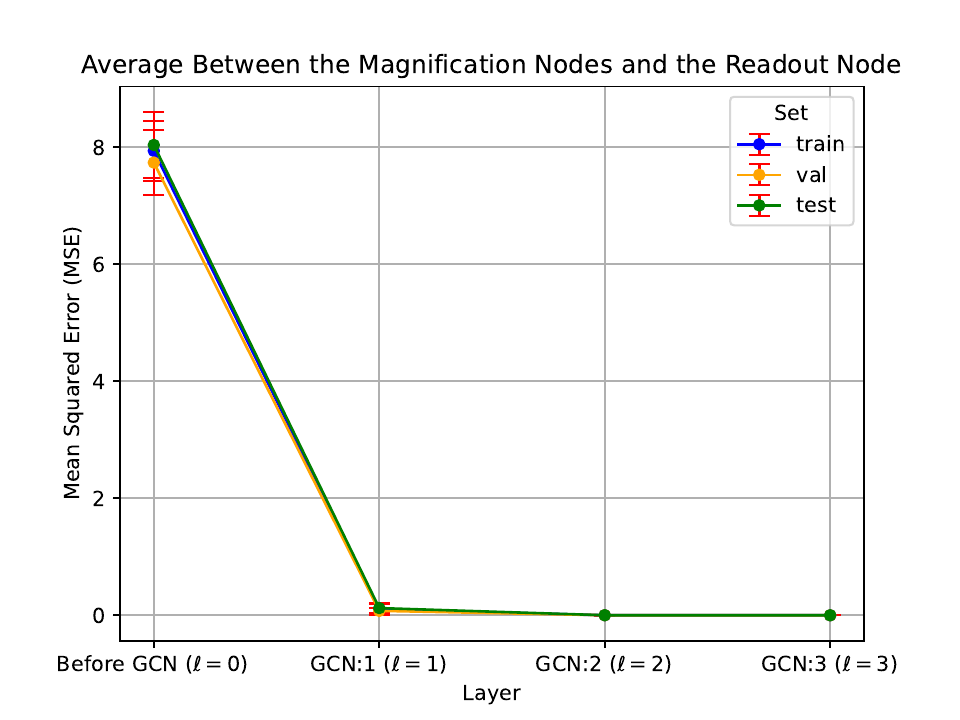}
    \caption{Empirical proof for intra-magnification convergence at $\ell = 3$.}
    \label{fig:gcn_convg}
\end{figure}

In addition to the theoretical analysis in Section \ref{sec:IMC}, we empirically demonstrate the intra-magnification convergence in Figure \ref{fig:gcn_convg}. In this experiment, we plot the mean squared error between all nodes in a magnification and the corresponding convergence node. As shown, the mean squared error at the third layer ($\ell = 3$) is nearly zero, providing empirical evidence for the convergence of the nodes, i.e., the nodes being pooled without the need for a pooling layer.
\label{ab:gcn}

\subsection{Ablation Study}
\subsubsection{Monte Carlo Test}
\label{monte-carlo-test}
As described in Algorithm \ref{alg:monte-carlo}, this test requires $m=400$. Therefore, we removed four slides that did not have 400 non-overlapping patches in $20x$ from the Bladder dataset and ran this test. Because of removing these four slides from the dataset, the result of Figure 4 in the paper is not comparable with any of the tables in the paper. In this test, a subset of the nodes is randomly dropped from each magnification layer (corresponding nodes in each magnification are removed), to create independent graphs. Then, training and inference happens and the results are reported at the end. 

\begin{algorithm}[h]
\caption{Monte Carlo Test on Graph Size}
\label{alg:monte-carlo}
\begin{algorithmic}[1]
\State $m \gets 400$
\State $step \gets 10$
\State \textbf{Load DATASET}
\State $D \gets \{\}$ \Comment{Node indices to be dropped}
\For{$\text{iter} \gets 1$ to $10$}
    \For{$G_r$ in DATASET}
        \For{$\text{count} \gets 10$ to $390$ step $step$}
            \State $D \gets \text{RANDOM}([1,\dots,m], \text{count})$
            \State $Q_{r,\text{count}} \gets G_r$
            \For{$\text{index}$ in $D$}
                \State $Q_{r,\text{count}} \gets \text{DROP}(Q_{r,\text{count}}, h_{\text{index}})$
                \State $Q_{r,\text{count}} \gets \text{DROP}(Q_{r,\text{count}}, h_{\text{index}}^{\prime})$
                \State $Q_{r,\text{count}} \gets \text{DROP}(Q_{r,\text{count}}, h_{\text{index}}^{\prime\prime})$
            \EndFor
            \State \text{STORE}$(Q_{r,\text{count}})$
        \EndFor
    \EndFor
\EndFor
\State \textbf{Use} $Q_{r,\text{count}}$ for cross-validation and inference
\State \textbf{Report:} Balanced Accuracy and Standard Deviation
\end{algorithmic}
\end{algorithm}

\begin{figure}
    \centering
    \includegraphics[width=7cm]{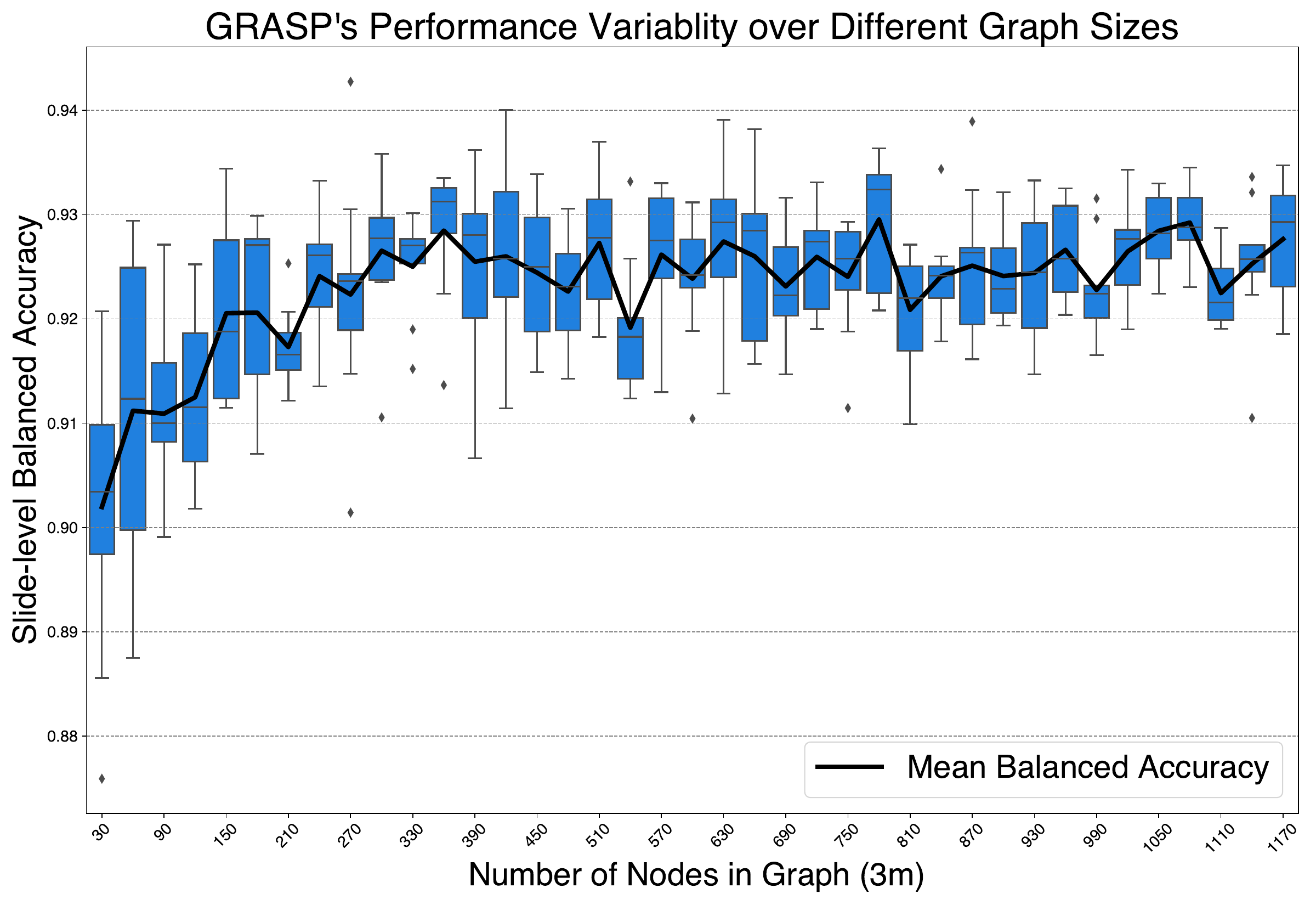}
    \caption{Monte Carlo experiments on the graph size. As the number of nodes increases, the uncertainty decreases and the model stabilizes.}
    \label{fig:monte-carlo}
\end{figure}
\begin{table}        
        \scriptsize
        \centering
        \caption{Average Performance on 3-folds and 10 random seeds based on KimiaNet's features.}
        \scalebox{0.99}{
        \begin{tabular}{c||cc}
        \toprule
        \multirow{2}{*}{}  & \multicolumn{2}{c}{\text{Bladder Cancer}}                         \\ \cline{2-3} \textbf{Model}     & Balanced Acc.        & F1 Score                 \\ \midrule
        \textit{Graph on} $\textbf{M}_1$        & $0.898 \pm 0.052$ & $0.890 \pm 0.047$\\
        \textit{Graph on} $\textbf{M}_2$        & $0.927 \pm 0.057$ & $0.928 \pm 0.051$ \\
        \textit{Graph on} $\textbf{M}_3$       & $0.905 \pm 0.035$ & $0.913 \pm 0.023$ \\
        \textit{Graph on} $\textbf{M}_1\&\textbf{M}_2$      & $0.919 \pm 0.032$  & $0.919 \pm 0.030$         \\
        \textit{Graph on} $\textbf{M}_1\&\textbf{M}_3$      & $0.917 \pm 0.031$  & $0.922 \pm 0.033$           \\
        \textit{Graph on} $\textbf{M}_2\&\textbf{M}_3$      & $0.926 \pm 0.024 $  & $0.934 \pm 0.022$          \\ \midrule
        \textbf{GRASP} (ours)    & $\mathbf{0.935} \pm \mathbf{0.011}$          & $\mathbf{0.937} \pm \mathbf{0.014}$      \\ \bottomrule    
        \end{tabular}
        
        }
        \label{table:ablation}
    \end{table}


In this experiment, we take a graph of size $3m=1200$, and randomly drop a set of its nodes, along with their multi-magnification correspondences, to build a new graph with a smaller size. For example, when $count=10$, we randomly drop 10 triplets of $(h_{index}, h_{index}^{\prime}, h_{index}^{\prime\prime})$ from the graph to create a new graph $Q_r$ of the size $1170$. Similarly, when $count=200$, we randomly drop 200 triplets of $(h_{index}, h_{index}^{\prime}, h_{index}^{\prime\prime})$ from the graph to create a new graph $Q_r$ of the size $600$. This is an aggressive way to create statistically independent graphs of smaller sizes. To capture statistical variance, we repeat the experiment $10$ times to create $40$ graphs with different sizes and report the model performance in Figure \ref{fig:monte-carlo}. To accomplish this, the same 3-fold cross-validation sets and $10$ random seeds have been used for all repetitions. Taking $10$ times repetitions of $40$ different graph sizes into account, we performed $12,000$ independent training and inference experiments. Algorithm \ref{alg:monte-carlo} demonstrates this experiment.

As can be seen in Figure \ref{fig:monte-carlo}, the performance of GRASP increases and stabilizes as the number of nodes increases. Since the standard deviation decreases as the number of nodes increases, it brings to light the concept of variance convergence, meaning that the model with $m \geq 200$ is fairly generalizable over different cross-validation folds and is statistically reliable in terms of performance. This is also in agreement with our theoretical expectation based on inter-magnification convergence that as $m$ grows, the model has better convergence resulting in more stability.


\subsubsection{Magnification Test}
\label{mul-mag}
To confirm that the idea of multi-magnification is valid and that multi-magnification is the cause for the model's performance, we design \textit{6} different experiments (repeated on 10 random seeds and 3 folds) on the Bladder dataset, with KimiaNet as the backbone, as our empirical evidence. These include evaluating the same model on only $\textbf{M}_1$, $\textbf{M}_2$, and $\textbf{M}_3$ fully connected graphs and on pairs of $\textbf{M}_1 \& \textbf{M}_2$, $\textbf{M}_1 \& \textbf{M}_3$, and $\textbf{M}_2 \& \textbf{M}_3$. The results in Table \ref{table:ablation} show that GRASP is superior to all other methods. One possible explanation is that for those single and paired graphs, three layers of GCNs most likely cause the aforementioned over-smoothing problem, which shows that GRASP can effectively capture the information contained in different magnifications and boost its performance.

\subsubsection{Graph Convolutions} 
\begin{figure}[h]
    \centering
    \includegraphics[width=8cm]{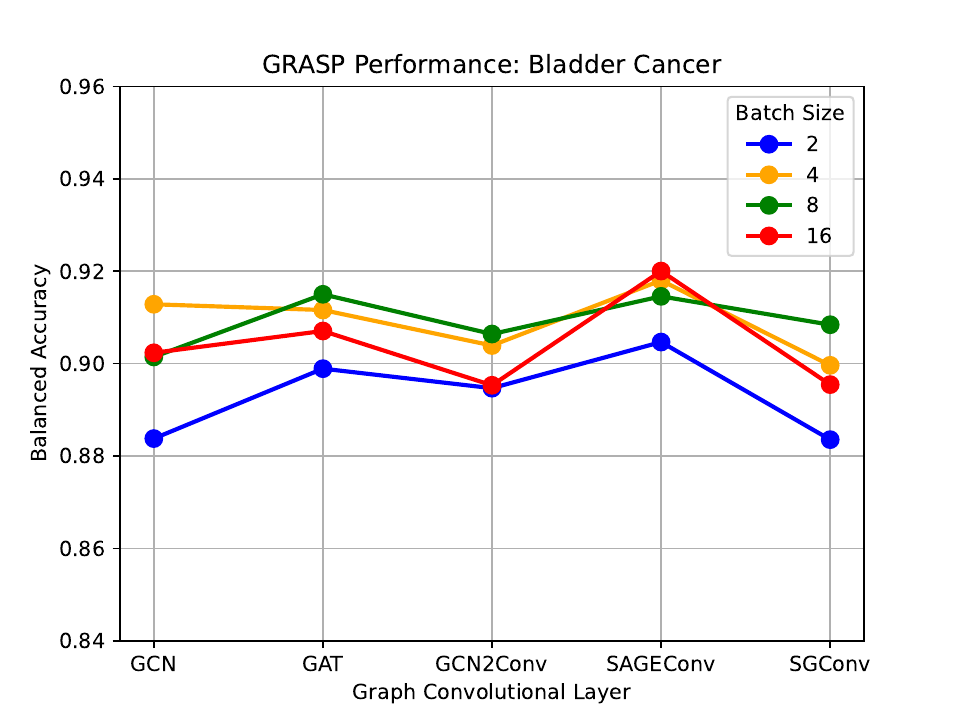}
    \caption{Ablation on the effect of different GNN structures benchmarked on 3 folds and 10 random seeds with different batch sizes. GNNs were taken from Deep Graph Library.}
    \label{fig:gcn_type}
\end{figure}

To study the impact of different graph convolutions on the performance of GRASP, we designed this experiment where we replaced the GCN layers in the original layers, with a newer version of graph convolutions. As seen in Figure \ref{fig:gcn_type}, GAT and SAGEConv improve the performance of the primary GCN-based GRASP in terms of Balanced Accuracy over different batch sizes. 
\label{ab:gcn}

\begin{table}[t]
\centering
\scriptsize
\caption{The average performance on 3 folds and 10 random seeds, based on Swin's features and the setting where all the patches were extracted from each WSI.}
\label{table:result_ab_gcn}
\large
\scalebox{0.65}{
\begin{tabular}{l|ccc}
\toprule
\multirow{2}{*}{Model} & \multicolumn{3}{c}{\large Bladder: Two Subtypes} \\ \cline{2-4} 
 & \multicolumn{1}{c}{Balanced Acc.} & \multicolumn{1}{c}{F1 Score} & \multicolumn{1}{c}{AUC}   \\ \midrule
\Large ZoomMIL &  $0.879 \pm 0.065$ &  $0.872 \pm 0.060$  & $0.951 \pm 0.031$    \\ 
\Large HiGT &  $0.720 \pm 0.049$ &  $0.658 \pm 0.042$  & $0.819 \pm 0.066$    \\
\Large H2MIL &  $0.877 \pm 0.050$ &  $0.871 \pm 0.035$  & $0.966 \pm 0.022$   \\ \midrule
\Large GRASP \small({GCN}) & $\bold{0.883 \pm 0.069}$ & $\bold{0.879 \pm 0.065}$ & $\bold{0.953 \pm 0.031}$ \\
\Large GRASP \small (GAT) & $\bold{0.917 \pm 0.013}$ & $\bold{0.907 \pm 0.017}$ & $\bold{0.978 \pm 0.007}$ \\ 
\Large GRASP \small (SAGEConv) & $\bold{0.936 \pm 0.023}$ & $\bold{0.932 \pm 0.015}$ & $\bold{0.988 \pm 0.008}$ \\ 
\bottomrule
\end{tabular}
}
\end{table}

\begin{figure}[!ht]
    \centering
    \includegraphics[width=8cm]{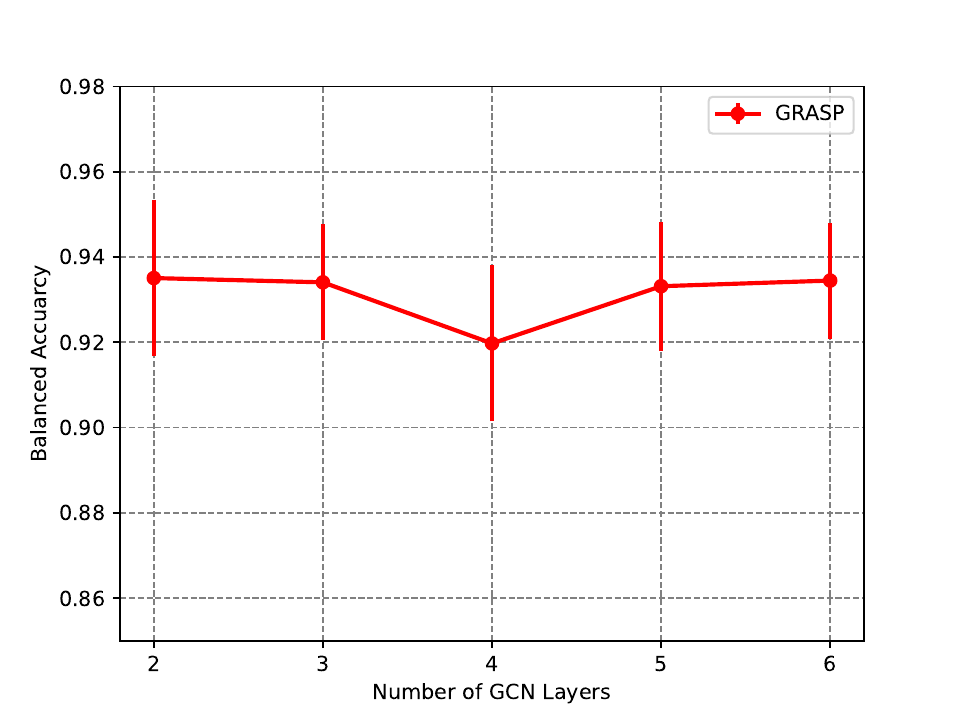}
    \caption{Our model's performance as the number of GCN layers increases. As can be seen, by increasing the number of layers from \textit{three} to \textit{four} a relatively large drop happens, showing that the model is being over-smoothed. However, the model is recovering this loss at $five$ layers or more yet with a relatively larger standard deviation compared to \textit{three} layers.}
    \label{fig:depth}
\end{figure}

\subsubsection{Patch Number}
\label{patch_num}
To compare the models' performance when all the patches from each slide are extracted, we designed this experiment and benchmarked the baseline against different variations of GRASP. GCN-based GRASP is superior to other multi-magnification methods such as ZoomMIL, H2MIL, and HiGT. Compared to Table \ref{table:result_ab_gcn}, the increased number of patches helps GRASP to improve, yet it degrades the performance of the other three models. However, single Magnification methods are more competitive with a higher density of patches. With this in mind, we followed the previous ablation study \ref{ab:gcn} and compared GRASP's performance with newer graph convolutions. Consequently, GRASP with SAGEConv outperforms all other models. This further emphasizes the flexibility of the graph structure that can easily employ different graph convolutions, which we belive is a unique advantage of GRASP.

\subsubsection{Graph Depth}
\label{sec:GD}
We experimented with GRASP with different numbers of GCN layers as shown in Figure \ref{fig:depth}. Firstly, \textit{three} layers of GCNs show the same performance as \textit{two} layers yet with lower standard deviation. Secondly, \textit{four} layers of GCNs show a sudden drop in performance and increase in standard deviation, which can be attributed to over smoothing problem \cite{oversmoothing, new_oversmooth}. In addition, with more than $five$ layers of GCNs, the network can recover the performance but with a slightly higher standard deviation and, clearly, an increased number of parameters. This shows that our original architecture of \textit{three} layers is the best choice in the trade-off of average accuracy and the model's reliability. Based on the discussion in \cite{new_oversmooth}, we expect the same phenomenon for different graph convolution types to happen.

\subsection{Graph-Based Visualization}
\label{sec:GBV}

Let's call the output of the classifier, $S$, where the logit for correctly classified slides/graphs is $S_c$. To visualize the importance of magnifications, we compute the magnitude of the gradient of a graph with respect to its node features at $l=0$ ($h_i^{(0)}$; for the sake of brevity, we drop the superscript $(0)$ form $h_i^{(0)}$ and show it as $h_i$), which is $\left|\frac{\partial S_c}{\partial h_i}\right|$ for the magnification $\textbf{M}_1$. $\left|\frac{\partial S_c}{\partial h_i}\right|$ is a vector of size $1024\times1$, and we have $m$ nodes giving result to $m$ such vectors. Likewise, we can define  $\left|\frac{\partial S_c}{\partial h_i^{\prime}}\right|$ and $\left|\frac{\partial S_c}{\partial h_i^{\prime\prime}}\right|$ for $\textbf{M}_2$ and $\textbf{M}_3$, respectively. Arranging these absolute gradients for each magnification in a matrix of size $m\times1024$ as follows,

\begin{equation}
    \label{eq:25}
    \textit{Heatmap}_{\mathbf{M}_1} =
    \left[
    \begin{array}{c}
        \left|\frac{\partial S_c}{\partial h_1}\right| \\
        \vdots \\
        \left|\frac{\partial S_c}{\partial h_m}\right|
    \end{array}
    \right]
\end{equation}

\begin{equation}
    \label{eq:26}
    \textit{Heatmap}_{\mathbf{M}_2} =
    \left[
    \begin{array}{c}
        \left|\frac{\partial S_c}{\partial h_1^{\prime}}\right| \\
        \vdots \\
        \left|\frac{\partial S_c}{\partial h_m^{\prime}}\right|
    \end{array}
    \right]
\end{equation}

\begin{equation}
    \label{eq:27}
    \textit{Heatmap}_{\mathbf{M}_3} =
    \left[
    \begin{array}{c}
        \left|\frac{\partial S_c}{\partial h_1^{\prime\prime}}\right| \\
        \vdots \\
        \left|\frac{\partial S_c}{\partial h_m^{\prime\prime}}\right|
    \end{array}
    \right]
\end{equation}

As such, putting matrices in \ref{eq:25}, \ref{eq:26}, and \ref{eq:27} together in a matrix gives us the overall heatmap for the graph, of the size $3m\times1024$, to compare the influence of each magnification:

\begin{equation}
    \label{eq:28}
    \textit{Heatmap} =
    \left[
    \begin{array}{c}
        \textit{Heatmap}_{\mathbf{M}_1} \\
        \textit{Heatmap}_{\mathbf{M}_2} \\
        \textit{Heatmap}_{\mathbf{M}_3}
    \end{array}
    \right]
\end{equation}

This is the heatmap depicted in Figure 9 as a graph-based heatmap, which shows how model focuses on different magnifications. 

Having the gradient for each node in the graph, we develop the concept of energy of gradients to find out which magnification(s) play a more important role in GRASP's final decision. To do so, we start by defining $\mathcal{E}_{\mathbf{M}_1}$ as follows,
\begin{equation}
    \mathcal{E}_{\mathbf{M}_1} = \sum_{i\in [1,...,m]} \left\| \frac{\partial S_c}{\partial h_i} \right\|_2^{2}
\end{equation}
similarly, for $\mathbf{M}_2$ and $\mathbf{M}_3$,
\begin{equation}
    \mathcal{E}_{\mathbf{M}_2} = \sum_{i\in [1,...,m]} \left\| \frac{\partial S_c}{\partial h_i^{\prime}} \right\|_2^{2}
\end{equation}
\begin{equation}
    \mathcal{E}_{\mathbf{M}_3} = \sum_{i\in [1,...,m]} \left\| \frac{\partial S_c}{\partial h_i^{\prime\prime}} \right\|_2^{2}
\end{equation}
Having these energies, the energy contribution of each magnification is calculated based on their relative share in the whole energy spent in the graph:
\begin{equation}
    {\mathbf{M}_1}\textit{'s contribution}= \frac{\mathcal{E}_{\mathbf{M}_1}}{\mathcal{E}_{\mathbf{M}_1}+\mathcal{E}_{\mathbf{M}_2}+\mathcal{E}_{\mathbf{M}_3}}
\end{equation}
\begin{equation}
    {\mathbf{M}_2}\textit{'s contribution}= \frac{\mathcal{E}_{\mathbf{M}_2}}{\mathcal{E}_{\mathbf{M}_1}+\mathcal{E}_{\mathbf{M}_2}+\mathcal{E}_{\mathbf{M}_3}}
\end{equation}
\begin{equation}
    {\mathbf{M}_3}\textit{'s contribution}= \frac{\mathcal{E}_{\mathbf{M}_3}}{\mathcal{E}_{\mathbf{M}_1}+\mathcal{E}_{\mathbf{M}_2}+\mathcal{E}_{\mathbf{M}_3}}
\end{equation}
Accordingly, the importance of each magnification can be quantified for further investigations. More samples are provides in Figure \ref{fig:histogram}.

\bibliography{lmrl2025_workshop}
\bibliographystyle{lmrl2025_workshop}

\end{document}